# AI Founding Fathers: A Case Study of GIS Search in Multi-Agent Pipelines

**Alvin Chauhan**
Independent Researcher
alvin.chauhan@gmail.com

**Abstract**

*Although Large Language Models (LLMs) show exceptional fluency, efforts persist to extract stronger reasoning capabilities from them. Drawing on search-based interpretations of LLM computation, this paper advances a systematic framework for understanding LLM reasoning and optimization. Namely, that enhancing reasoning is best achieved by structuring a multi-agent pipeline to ensure a traversal of the search space in a gradual, incremental, and sequential (GIS) manner. Stated succinctly, high-quality reasoning is a controlled, incremental search. To test this framework, we investigate the efficacy of recursive refinement (RR)—an iterative process of self-criticism, adversarial stress-testing, and integrating critical feedback—as a practical method for implementing GIS search. We designed an experiment comparing a simple, linear pipeline against a complex, explicitly structured pipeline leveraging a recursive refinement layer. The multi-agent models were constructed to reflect the historical personas of three US Founding Fathers (Hamilton, Jefferson, and Madison) using RAG-powered corpora and were prompted to generate responses to three contemporary political issues. Model performance was evaluated using a two-tiered approach: a quantitative score from an LLM arbiter agent and qualitative human judgment. Our results revealed that the complex model consistently outperformed the simple model across all nine test cases with an average arbiter-outputted score of 88.3 versus 71.7. The complex model's arguments were superior in analytical depth, structural nuance, and strategic framing. We conclude that recursive refinement is a robust architectural feature for enhancing LLM reasoning via GIS search.*

## 1.0 Introduction

Large Language Models (LLMs) increasingly resemble digital alchemy: they are the nerve center of computational solutions for an expansive and seemingly unlimited range of problems. Even hardened skeptics concede their marked ability to dispose of boilerplate and formulaic tasks. With their powerful generative capabilities, these models show exceptional levels of fluency and performance across various domains where progress was previously haphazard and uneven.

Despite their impressive feats, LLMs are not infallible instruments. Persistent issues surrounding factual accuracy, contextual understanding, and handling of prompt instructions continue to hamstring applications reliant on LLMs. Various techniques however have emerged to remedy these flaws and place language models on a more rigorous foundation. This analysis gauges the efficacy of structuring a multi-agent pipeline according to a GIS-search architecture through the application of recursive refinement – the use of self-criticism, stress-testing, and integrating critical feedback – while harnessing the simulated wisdom of America's premier political theorists: the US Founding Fathers.

Studying language model performance, researchers found that instructing an agent to perform its analysis or task in a discrete, sequential series of steps improved output quality. Work in this area generated the concept of Chain-of-Thought reasoning as well as prompt engineering practices that enhanced in-context learning. Extending these ideas, this paper advances a systematic framework for understanding LLM optimization, namely, that improving model performance requires structuring a multi-agent pipeline to ensure a traversal of the search space that is gradual, incremental, and sequential. Using this lens, the value of recursive refinement resides in its ability to guide search traversal in a more controlled and constrained manner.

To test this thesis, we implemented two different multi-agent models: a linear pipeline without a recursive refinement layer (the 4-agent simple model) and an explicitly structured pipeline with a recursive refinement layer (the 8-agent complex model). The complex model was designed to perform a repertoire of self-critical tasks including:

- Identifying flaws and weaknesses in intermediate arguments from the perspective of different interlocutors and based on alignment with internal standards of self-consistency.
- Anticipating counterarguments from adversaries and developing counter-responses to these criticisms by leveraging different rhetorical strategies.
- Applying a multi-tiered evaluation criteria to assess, select, and integrate the most effective responses into a single, convincing final argument.

This experimental setup was executed using a simulated panel of three US Founding Fathers – Alexander Hamilton, Thomas Jefferson, and James Madison – and was designed to elicit each historical persona's response to three contemporary political issues. Model performance was compared and evaluated according to both an LLM arbiter agent and qualitative human judgment.

The selection of the US Founding Fathers as the interlocutors of this simulation had several motivations. It served as a test of an LLM's ability to create authentic historical personas capable of reasoning about contemporary political issues. More pointedly, the founding fathers' rich intellectual and polemical tradition of discussion, criticism, debate, and dissent provided a striking conceptual parallel to the complex model's adversarial recursive design. The simulation ultimately provides a broader view into whether a framework of ideas and principles forged in a pre-modern era can still appeal to us and influence our thinking today.

Ultimately, this paper shows that a structured multi-agent pipeline leveraging recursive refinement enhances language model performance. Imposing a structured and iterative process of self-criticism, adversarial stress testing, and incorporating critical feedback strengthens the quality of reasoning and argumentation. Our framework of controlled and sequential search more broadly explains the power of prompting and in-context learning, and provides insight into the mechanics and causality of recursive refinement.

## 2.0 System Architecture and Design

The design of the founding fathers AI simulated panel was informed by separation of concern principles focused on modularity, decoupling, and decomposition. This architecture enables the decomposition of a complex multi-agent framework into a series of distinct, linear, and robust components.

### 2.1 The Agent Pipeline Architecture

The core system consists of two sequential multi-agent pipeline architectures. The output of one specialized agent serves as the input for the subsequent agent in the pipeline thus supporting a unidirectional flow of information. This enforced linearity was a deliberate design choice to model the natural progression of developing and refining an argument. This structure further provides a clear "assembly line" for thought and allows for the inspection of intermediate outputs at each stage of the overall argument generation process. Chaining specialized agents in this manner aligns with best practices in prompt engineering and supports the decomposition of complex tasks into a series of simpler, sequential prompts.

### 2.2 Key Software Engineering Patterns

To ensure the system was modular, maintainable, and flexible, several key software engineering patterns were implemented.

<u>Separation of Concerns:</u> Each agent is a distinct class with a single, well-defined responsibility. For example, the Researcher Agent is tasked only with querying the knowledge base, while the Communicator Agent is responsible only for revising and polishing the prose. This separation makes the system easier to develop, debug, and modify. If one agent's logic needs to be updated, the others remain entirely unaffected.

<u>Dependency Injection:</u> The design of the system avoids having agents create their own dependencies. Instead, high-level components like the RAG system and the prompt dictionary are created once in a master script and are then "injected" into the agents that use them. This decouples the agent's logic from the specific tools it uses. This design offers great flexibility. For instance, the underlying vector database can be swapped out by changing only one line in the main script, with no modifications needed for any of the agent classes. This also makes the system highly testable as test dependencies can be injected to test agents in isolation.

<u>Externalized Configuration:</u> All agent instructions are decoupled from the Python logic and managed in a central file: prompt.yaml. This separates the agent's "brain" (the prompt) from its "body" (the Python code) and enables rapid iteration and experimentation. Leveraging this, an agent's strategy, logic, or design can be significantly changed simply by editing a text file without having to apply changes to the underlying codebase.

"Reason-Then-Extract" Pattern: To enforce the JSON-only rule and ensure pipeline stability, the base agent class employs a robust two-step process. The first LLM call performs the complex reasoning, and a second specialized LLM call is used to reliably extract the clean JSON object from the first call's potentially convoluted output. This small architectural choice dramatically increases the reliability of the entire system.

## 2.3 The RAG (Retrieval-Augmented Generation)

The system was built atop a RAG architecture to ensure all arguments were grounded in historical facts and would authentically replicate the founders' personas. It leverages the following:

- Database Vectorization: This was implemented using the Sentence-Transformers library to generate text embeddings and ChromaDB as the vector database to store and retrieve them.
- Persona-Specific Retrieval: The key feature of the RAG system is its use of metadata filtering. Each piece of text from the corpora is tagged with its author's name (e.g. "Hamilton"). When the Researcher Agent queries the database, it filters by this tag to ensure that a query for Hamilton's views only returns results from his own writings and prevents factual conflation across different personas.

## 2.4 Advanced Prompt and Agent Design

The system's performance relies on a sophisticated approach to prompt and agent design leveraging the following mechanics:

- Chain-of-Thought (CoT): Most of the prompts deployed instruct the LLM to first reason through its task in a <thinking> block before providing a final answer. This forces a more robust, step-by-step reasoning process leading to higher-quality outputs.
- Structured I/O: All prompts use XML tags (e.g., <instructions>) to clearly delineate different parts of the input. Critically, all agents are required to return their output as a single, valid JSON object.

## 3.0 Experimental Framework

### 3.1 Agent Design and Structure

Understanding the cognitive functions that animate human reasoning, planning, strategizing, and language strikes at the very core of scientific and philosophical inquiry. Within the world of language models, prompt engineering has emerged as one of the central arenas in which such exploration can occur. Formulating the right prompt with the right instructions provides a mechanism through which we can study and test theories of how "reasoning" occurs within a computational framework.

The experimental framework was designed to directly test our GIS thesis by comparing two distinct pipeline architectures:

- The 4-Agent Simple Model (Figure 1) serves as our baseline control. It represents a linear reasoning process without the explicit gradual, incremental, and sequential properties of the GIS-search framework.
- The 8-Agent Complex Model (Figure 2) serves as our explicitly structured, GIS-search informed pipeline. It implements GIS search through a 4-step recursive refinement layer to enforce a more controlled traversal of the search space.

The agents that comprise these two pipelines are described in detail below.

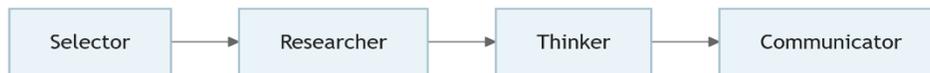

*Figure 1. The 4-Agent Simple Model.*

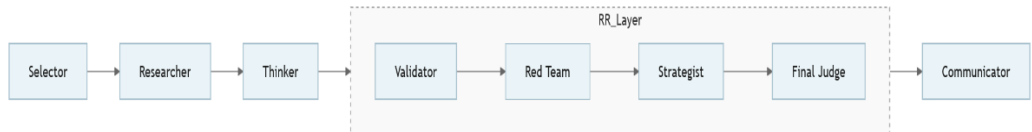

*Figure 2. The 8-Agent Complex Model.*

The Selector Agent

The Selector agent initiates the multi-agent pipeline. It deconstructs the contemporary political question into a strategic blueprint from which the founding father persona can develop an argument rooted in a particular worldview.

Receiving the debate topic and the founder's persona profile as input, the agent performs a three-step analysis:

- It selects a single core principle from the founder's philosophy that is most relevant to the topic.
- It identifies a historical precedent where the founder applied that same principle.
- It finds an ideological ally—a historical thinker who was aligned with the founder on that specific principle.

The agent's output is a structured JSON object containing these three elements. This object serves as the foundational brief providing the essential building blocks for the Researcher Agent and Thinker Agent which appear later in the pipeline.

### The Researcher Agent

The Researcher Agent grounds the historical persona's output in a detailed knowledge base of curated and representative works authored by the founding father to aid in authentically and accurately capturing the founding father's voice.

Unlike the other agents in the pipeline, the Researcher Agent does not rely on an LLM for its reasoning. Its sole function is to perform a targeted query in the RAG system's vector database. Receiving the strategic brief from the Selector Agent, it develops several precise queries based on the identified core principle, historical precedent, and ideological ally outputs. A critical feature of this process is the use of metadata filters which ensures that queries for a specific founder's views only retrieve information from that founder's own writings.

The agent's output is a "research dossier" containing the most relevant text passages from the historical corpora. This dossier provides the necessary factual grounding and direct evidence for the subsequent Thinker Agent to use when constructing its arguments.

### The Thinker Agent

The Thinker Agent is the core generative component of the system. It is responsible for synthesizing and transforming the strategic framework and research materials into fully developed arguments. It receives the outputs from the Selector and Researcher agents and is instructed to follow a strict prioritization hierarchy to ensure that its reasoning is grounded primarily in factual evidence and the founder's core principles.

The role of the Thinker Agent differs between the two models. In the simple model, the Thinker Agent is prompted to produce a single, direct, and evidence-based argument. This serves as the baseline generative output for the experiment.

In the complex model, in contrast, the Thinker Agent is tasked with generating three distinct types of argument: an orthodox, an unorthodox, and a pragmatic option. This process of creating multiple, varied lines of reasoning provides a rich set of candidate arguments for the recursive refinement layers to analyze and improve upon.

The agent's final output is a JSON object containing either the single argument or the three distinct arguments depending on the model being run.

### The Validator Agent

The Validator Agent is the first stage of the recursive refinement architecture. Its primary function is to act as an objective internal critic and systematically evaluate the three candidate arguments generated by the Thinker Agent to identify the strongest and most compelling one.

To accomplish this, the agent employs a detailed, weighted scoring rubric. It assesses each argument against three criteria: principles consistency (60%), personality consistency (25%), and intellectual strength (15%). After scoring each candidate, it calculates a final composite score and selects the single argument with the highest score.

The Validator Agent's output is a JSON object containing the full evaluation results. It passes the text of this single winning argument to the next stage of the pipeline where it is subjected to adversarial stress-testing by the Red Team Agent.

### The Red Team Agent

The Red Team Agent is the second stage of the recursive refinement architecture. Its sole purpose is to act as an adversarial counterpart to stress-test the winning argument from the Validator Agent to find its single most critical vulnerability.

To achieve this, the agent is given a purely adversarial persona and a precise, three-step analysis process. First, it identifies all potential "attack vectors" including both internal logical flaws and powerful external counterarguments from the perspective of the other founders. Second, it simulates how the original founder would likely defend against each of these attacks. Finally, it selects the single vulnerability that was the most difficult to defend against - the most damaging and potent liability.

The agent's output is a JSON object that identifies and describes this single critical vulnerability which is then passed to the Strategist Agent.

### The Strategist Agent

The Strategist Agent is the third stage of the recursive refinement architecture. After the Red Team Agent identifies a key vulnerability, the Strategist Agent develops a range of defensive rhetorical strategies.

Receiving the critical vulnerability as its primary input, the agent is tasked with generating three distinct counterresponses each representing a different rhetorical approach:

- A direct rebuttal that directly addresses and challenges the criticism head-on.
- A reframe and minimize response that diminishes the criticism's importance and significance.
- A concede and outweigh response that argues the flaw is an acceptable trade-off given the net benefits accrued.

The agent's output is a JSON object containing these three distinct strategic responses. This provides the subsequent Final Judge Agent with a range of options to choose from for strengthening and finalizing the argument.

### The Final Judge Agent

The Final Judge Agent is the fourth and final stage of the recursive refinement architecture. Its role is to act as the ultimate arbiter evaluating the range of rhetorical options and to select and integrate the best one to produce a single, unified, cohesive, and robust argument.

The agent receives the original argument and the three strategic counterresponses from the Strategist Agent. It then performs a three-step decision-making process. It first constructs four candidate arguments: the original version and three new versions where each version integrates one of the generated strategic responses. It then evaluates all four candidate arguments against the primary criteria of persuasiveness and resilience to criticism. Finally, it selects the single winning argument.

Its output is a JSON object containing the final, integrated text of the winning argument along with a justification for its choice. This revised and refined argument is then passed to the Communicator Agent for stylistic finishing.

### The Communicator Agent

The Communicator Agent is the final stage in both the simple and complex pipelines. Its function is to act as a skilled ghostwriter able to translate the final, logically structured argument into a powerful and persuasive written statement delivered in the founder father's authentic voice.

The Communicator Agent receives the final argument brief—either directly from the Thinker Agent in the simple model or from the Final Judge Agent in the complex model. Its primary task is not to construct new arguments or ideas, but to perform a stylistic enhancement. It analyzes the founder's persona profile, focusing on communication style and representative prose, and then rewrites the argument to match the specific tone, vocabulary, and rhetorical profile of the historical figure.

The agent's output is a JSON object containing the final, polished statement which represents the historical agent's complete response in the simulation.

### The Arbiter Agent

The Arbiter Agent is the final component of the experimental framework and acts as an impartial judge providing a definitive answer to the project's core research question. Unlike the other agents, it does not adopt a historical persona and operates outside the main simulation pipeline.

Its sole function is to conduct a direct comparison between the final arguments generated by the simple and complex models for a given founder. The agent is provided with both arguments and a detailed 4-part scoring rubric with equally weighted criteria including structure, depth, support and justification, and rhetoric and style. It systematically scores both arguments against this rubric and provides a quantitative score and a qualitative justification for its decision.

The agent's output is a JSON object that declares the winning model and provides a detailed justification for its verdict based on the specific strengths and weaknesses identified during its analysis.

### 3.2 Simple Versus Complex Model

To answer the core research question, we ran the simulation by posing three distinct contemporary questions to each of the three founding father personas (Alexander Hamilton, Thomas Jefferson, and James Madison). Each of the nine simulated scenarios was run through both the simple and complex models.

The final outputs from both models were then subjected to a two-layer evaluation: a quantitative judgment rendered by the Arbiter Agent based on its scoring rubric and a qualitative human judgment to reach a final decision.

## 4.0 Results and Analysis

Across all questions and all founding father agents, the complex model outperformed the simple model in its responses as measured both by the Arbiter Agent and qualitative human judgment. Our results support the hypothesis that recursive refinement improves the reasoning capabilities of language models. Although all formulated responses were substantive and coherent, the two Jefferson-agent answers on immigration (question two) suffered from a misreading of the nomenclature and misinterpretation of the context underlying the question statement. This gave rise to spurious assertions which strained the overall argument.

The quantitative results from the Arbiter Agent's scorecard were definitive: the complex 8-agent model outperformed the simple 4-agent model in all 9 comparative scenarios. As shown in Figure 3, the overall average complex model score was a full 16.7 points higher than the simple model's overall average score (88.3 vs 71.6). This performance gap was consistent across all personas with the complex model scoring higher for Hamilton (87.5 vs. 71.7), Jefferson (87.5 vs. 66.7), and Madison (90.0 vs 76.7) as shown in Figure 4.

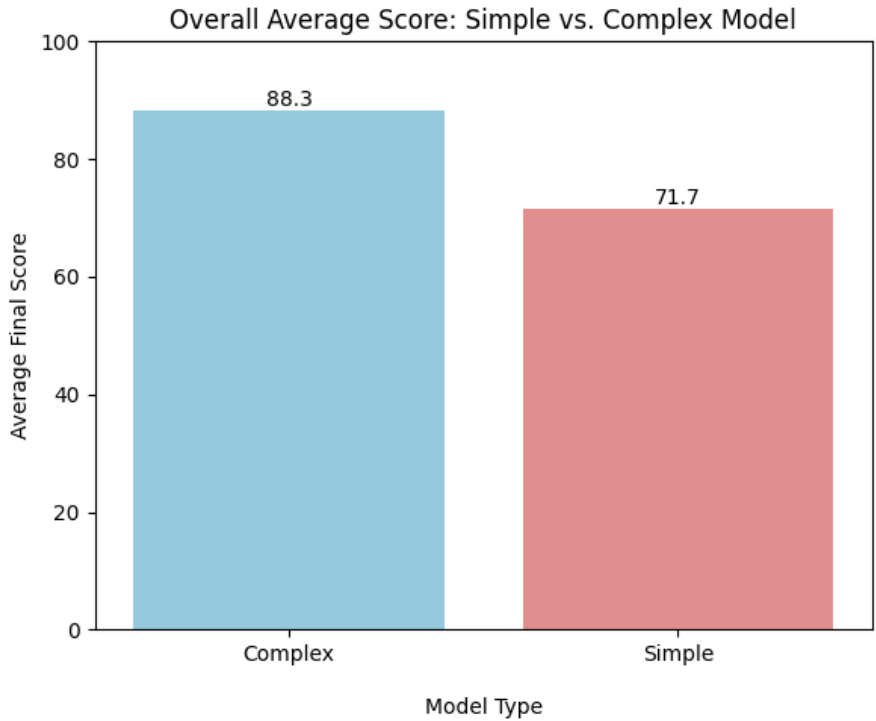

*Figure 3. Overall Average Score: Simple vs. Complex Model.*

Given the centrality of creating authentic historical personas to the experiment, it is worth considering how well each model performed in faithfully replicating the founding father's voices. While the arbiter consistently rated the complex model outputs as more rhetorically compelling, assessing authorial style requires going beyond mere eloquence and expression.

Recognizing the inherent subjectivity, complexity, and expertise that such a task requires – calling on historical knowledge, linguistic sophistication, and a strong familiarity with the speaker's prose style – selecting a winner is fraught with uncertainty and imprecision. That this is not a clear-cut decision is testament to the strength of the baseline model architecture to develop a prose style in the absence of recursive refinement. While it could be argued that the heightened sophistication and nuance of the complex model positions it closer to the elevated discourse of the founding fathers, both model responses are persuasive, credible, and bear the signature literary characteristics evocative of the founding fathers.

The analysis below captures the primary distinctions between the responses of the two models. Note that our work deliberately omits consideration of rhetorical components like rebuttals, refutation and anticipation of criticism given that this was an intentional design feature that separated the two models.

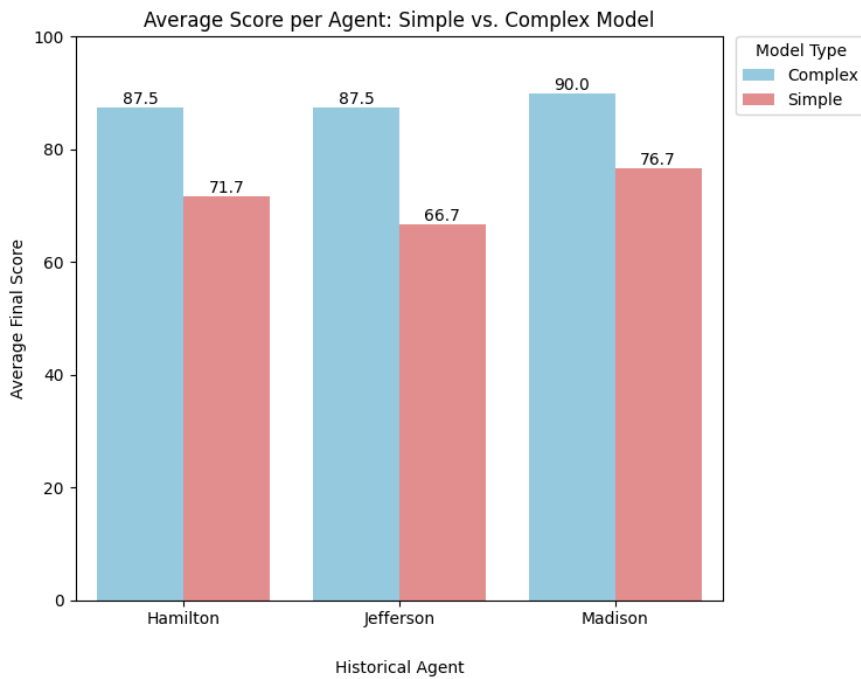

*Figure 4. Average Score Comparison by Historical Agent.*

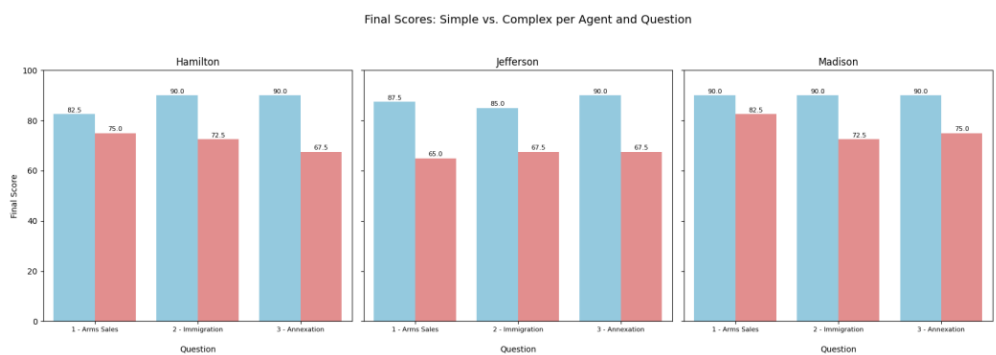

*Figure 5. Detailed Final Scores by Agent, Question, and Model Type.*

## 4.1 From Absolutist and Monolithic to Qualified and Nuanced

The simple multi-agent model tends to produce responses that are sweeping and unqualified in their pronouncements – often developing an absolutist, monolithic position which it then forcefully and continuously asserts. The complex model on the other hand offers a more qualified, balanced appraisal. It introduces caveats and shuns ideological extremes, while building an argument with greater depth, nuance, and trade-offs.

For instance, the simple Hamilton agent characterizes opposition to arms sales as tantamount to "treason to the national interest" and commandingly invokes the "preservation of the union" as a key consideration. Complex Hamilton, in contrast, admits that statecraft is the "not the art of avoiding all risks – it is the art of choosing which risks serve the national interest" and acknowledges the broader dangers involved while denying that "those dangers outweigh the fundamental imperative of industrial development". The complex mode accounts for trade-offs and risks in a way that the simple model does not.

In discussing immigration, the complex Madison agent similarly establishes a moderate position declaring that "the prudent course, therefore, lies neither in the extreme of borders thrown open without discrimination ... but rather in a calibrated approach that serves multiple constitutional ends simultaneously while managing, though not eliminating, the acknowledged risks", a sentiment absent in the response of its simple counterpart which instead focuses on serving the interests of citizens and social cohesion.

Along the same lines, the complex Hamilton agent on the topic of immigration stresses that "the government must retain the flexibility to adjust immigration flows based on economic conditions, strategic needs, and the capacity of newcomers to assimilate into our national project", whereas the simple Hamilton single-mindedly focuses on prioritizing stability and the interests of existing citizens omitting consideration of evolving circumstances or contingencies. The complex Hamilton in contrast strikes a temperate and pragmatic tone: "What matters is not adherence to any ideological extreme—neither the cosmopolitan fantasy of open borders nor the timid defensiveness of excessive restriction—but rather the instrumental use of immigration policy to build national strength."

### 4.2 Multidimensional Enrichment and Connection to Broader Themes

Whereas the simple model develops a unified yet unitary argument, the complex model develops its reasoning across multiple dimensions and connects its claims to broader and more robust themes and principles. The simple Hamilton agent on arms sales cites national power, but its complex counterpart touches on "industrial supremacy and manufacturing capacity" and analyzes the broader implications of selling arms: "advanced weaponry strengthens our industrial base, perfects our manufacturing processes, creates employment for our citizens, and establishes dependencies that translate into diplomatic leverage". Simple Madison on immigration invokes the Federalist papers and calls for social cohesion, but complex Madison – when invoking the Federalist papers – incorporates the political doctrine of factionalism and the need to balance competing societal segments and interests to prevent coercion and preserve stability.

### 4.3 Expansion and Further Elaboration

The complex model regularly expands and further develops ideas found in the outputs of the simple model. On the question of annexation, simple Hamilton says, "prudence demands we seize opportunities for peaceful expansion that enhance our commercial strength" and suggests, "incorporation of neighboring territories must proceed through negotiation and mutual advantage". Complex Hamilton however offers a systematic and tactical implementation of the idea of "continental economic integration" by developing a four-stage blueprint for achieving this through commercial treaties, infrastructure integration, political unification, and defensive alliances. The complex model grasps the need to devise a plan with the granularity and structure to appeal to others and eschews wholesale declarations.

### 4.4 Jefferson Goes Astray on Immigration?

On the question of immigration, both the simple and complex Jefferson agents appear to go astray in their understanding of the question. They both conflate selection of high-skilled immigrants with a covert effort to sustain aristocratic and hereditary privilege. The Jefferson agents see this immigration policy as tantamount to the imposition of arbitrary and capricious standards by an unaccountable centralized authority. Rather than furthering meritocracy, this policy is seen as anti-meritocratic with the ultimate consequence of inadvertently replicating the hierarchical societies of Europe. It is still worth pointing out however that complex Jefferson still outperformed its simple counterpart when judging performance according to both Arbiter score and human judgement despite the shared conceptual divergence.

Notably, complex Jefferson even goes so far as to denounce an incipient "aristocracy of talent" in its criticism. This sharply contradicts the views of the actual Jefferson who supported the creation of a "natural aristocracy" – a class of elites who secured legitimacy through talent and merit instead of hereditary privilege. Instead, the Jefferson agents regarded vetting of immigrants as a form of maintaining and reinforcing privilege.

Setting aside for a moment this characterization of the model as going astray, are there perhaps elements of a more subtle critique lurking amid the models' responses? The spirit of the question was to invite debate on cultivating an immigration policy prioritizing high-skilled workers in a decidedly pro-meritocratic manner. Yet this very premise is open to challenge: can the state really set standards that will identify talent or will these requirements necessarily be ad hoc, ill-conceived, and inadequately designed? Will they inevitably degenerate into a slew of arbitrary and idiosyncratic demands subject to abuse by the state?

While it is possible that the Jefferson-agent is surfacing such criticisms, it is also possible that the model is operating under the pervasive influence of Jefferson's deep-seated tendency toward republicanism and egalitarianism (expressed through the model weights) and that this invariably colors all its responses in this domain. This divergence illustrates the powerful influence of latent statistical tendencies in blunting and overriding more pointed contextual prompts.

## 5.0 Discussion

Our findings suggest that structuring a multi-agent pipeline according to a GIS-search architecture creates more robust and persuasive lines of argument. Integrating layers of critique and feedback into a multi-agent pipeline consistently produces stronger levels of model performance. Deconstructing this result to isolate causality however calls for deeper study of the interaction of in-context learning, computational paradigms, and model architecture. This investigation begins with an exploration of reasoning capabilities and builds toward the broader project of rationalizing model behavior, identifying the drivers of learning and performance, and imposing structure on complexity.

## 5.1 Model Reasoning – Empirical Advances and Theoretical Pitfalls

Recursive refinement and reasoning are inextricably linked. Refinement, after all, is an advanced form of reasoning itself – revision directly alters the logic, content, and arrangement of ideas. Dramatic advances in this space flow from research and innovation in prompting strategies. A closer examination of prompting thus lays the groundwork for a more rigorous examination of recursive refinement and GIS-search and how it amplifies performance.

Research on prompting has yielded several notable techniques for improving LLM reasoning capabilities. The methods applied and tested vary in design and complexity. (Kojima et al., 2022) found that merely adding "Let's think step by step" before each answer significantly improved accuracy. (Wei et al., 2022) unlocked further performance gains with Chain-of-Thought (CoT) prompting and the inclusion of intermediate reasoning steps before an answer. And a similar approach by (Zhou et al., 2023) decomposed a given problem into a sequence of sub-problems with later sub-problems leveraging solutions to earlier ones.

In explaining the phenomenon of "in-context learning", investigators frequently resort to anthropomorphic metaphors and analogies. In the same way that humans – so we are told – benefit from breaking down a complex problem into smaller problems and simpler steps, so too do language models learn from comparable heuristics.

Scrutinized more closely, however, these parallels are soon found to be flawed and without foundation. Several lines of research show that adversarial prompting via the inclusion of irrelevant, misleading, or counterfactual information – adjustments which would be expected to destroy the reasoning chain and render the outputs meaningless or erroneous – instead leaves the outputs almost unchanged (Madaan et al., 2023; Webson & Pavlick, 2022). Undercutting prevailing reasoning narratives, (Zhao et al., 2021) showed that changing the order of training examples caused model accuracy to significantly vary. This sensitivity to ordering was further confirmed by (Lu et al., 2022).

Without even grappling with the existing opacity and inscrutability of human reasoning, it is apparent that using a human-oriented cognitive lens to explain model reasoning leaves much to be desired. We must instead search for a better interpretation mechanism, one that is afforded to us by the concept of search space optimization.

## 5.2 Prompting as Search Optimization: Gradual, Incremental, Sequential Searching

Although efforts to wrangle with language explainability still lack precision and definitiveness, it is widely understood that machine learning models generate fluent responses through exploitation of patterns, semantic relationships, and statistical associations. Deep neural networks and advanced architectures like transformers powerfully harness this ability. They do so through computational depth and algorithmic complexity – imbuing models with qualities like context, similarity, relative importance, and positional information. These enrich pattern recognition and enable the identification of complex and multidimensional relationships and connections.

The dense, opaque, and intractable complexity of such machinery precludes formulaic description and reductive explanations. Efforts at interpretation are best mediated through a computational framework treating model execution as the traversal of a search space. In-context learning can now be framed as search space optimization. This paper lays out the view that enhancing reasoning and in-context learning, and consequently, optimizing model performance, calls for structuring a multi-agent pipeline to ensure a traversal of the search space that is gradual, incremental, and sequential (GIS).

Existing prompting techniques like CoT reasoning and problem decomposition, as well as architectural features like multi-agent specialization, can all be seen as methods for inducing a more gradual, incremental, and sequential search. This paradigm allows us to rationalize core findings from the literature, make sense of counterfactual results, and prepare the ground for exploring other optimization methods.

A justification for this prescriptive guidance is warranted. For instance, why aren't LLMs able to generate a correct answer on the first attempt with just a succinct yet complete prompt? The vast dimensionality of the search space provides an immediate answer: we may intuitively regard the probability of the model succeeding on the first shot with limited context to be rather small given the massive density of the search space.

This is consistent with the observations of (Zhao et al., 2021) and (Holtzman et al., 2022) who showed that small changes to the prompt resulted in large performance variations. Outputs are extremely sensitive to changes in inputs. Prompt engineering thus helps guide this volatile search in a more accurate direction by grounding the computation in contextually rich information and allowing for stronger recognition of patterns and relationships that can more clearly define a path through the search space.

Classic prompting techniques illustrate this idea. Instructing the model to "think step by step" constrains it from making early sweeping, dramatic shifts to the conjectured optimal location, and instead – through decomposition and supporting contextual information – encourages incremental and smaller steps toward a final location.

Intermediate steps thus provide more controlled steering and course correction for a traversal of the search space.

With CoT prompting, the concept of a sequential flow of reasoning more clearly emerges. A sequence of steps can be established with successive steps leveraging the work and solutions of earlier steps. In this manner, a solution is gradually and progressively constructed. Conceptually, this aligns with our intuitive notion that beginning our search in an area far from the optimal location makes it far harder to ultimately end up there. The accumulation of small movements, even if some are individually skewed, can still bring us to the desired search space spot with greater likelihood if they are correct on aggregate.

While other interpretations of prompting have surfaced, none survive critical scrutiny. For instance, (Wei et al., 2022) tested whether CoT prompting was only useful as a means of activating relevant knowledge during pretraining. They analyzed the results of an alternative arrangement where the CoT prompt was provided after the answer with the aim of breaking the "chain of reasoning". They found that this performed no better than the baseline model. Having the chain of reasoning serve as a precursor to the answer clearly enhanced in-context learning. Stated otherwise, steering the search in a sequentially logical manner pays dividends. In a similar vein, adopting a RAG architecture can be seen as a means of anchoring the model in the correct overall vicinity before more targeted, incremental search fine-tuning can occur.

This similarly provides a basis for understanding the success of "least-to-most" prompting used by (Zhou et al., 2023). It also explains why the random arrangement of examples in a prompt causes performance to significantly deteriorate. The lack of sequential progression damages the ability for effective search to take place.

The more striking example that counterfactual prompting still leads to reasonably accurate results is on the surface more puzzling, but can still be rationalized. By taking gradual, incremental steps, the "noisy" signals are limited in magnitude, and lack the cumulative strength to influence the overall path progression. The content and structure of the prompt supply enough "correct" anchoring to activate the correct relationship and pattern matching to enable accurate searching.

Searching does not imply a single path – multiple searches are possible, and the same answer can be reached in different ways. This is explored in the work of (Wang et al., 2023). Sampling from the language model's decoder, they contemplate the diversity of reasoning paths the model might follow. Noting the universal truth expressed in the aphorism, "there are several ways to attack a problem", they find that correct reasoning processes show greater alignment and consistency than incorrect ones. This strategy of choosing the reasoning path with the greatest level of self-consistency led to a significant increase in model performance. (Wang et al., 2023) argue that this overcomes the limitation of "greedy decoding" and helps break out of restrictive local optima.

## 5.3 Recursive Refinement: An Enhanced Search

With our framework for reasoning in place, we now turn to the anatomy of refinement. In our earlier commentary, we sowed the seeds for this undertaking: given the vast and complex multidimensionality of the search space, it is unreasonable to assume that the model can arrive at an optimal solution on its first attempt. This is not to say that the model's output is wildly inaccurate. Recall that it has already engaged in intense processing and computation and is likely already in a relatively good location. Refinement, however, with its connection to targeted revision rather than wholesale reconstruction – allows for further tuning and adjustment. Refinement therefore allows the model to move to another point in its neighborhood that is more optimal.

Our experimental findings provide support for this search-based framework. While the simple model outputs were generally quite solid and substantive, the ability of the complex model to calibrate, adjust, and fine-tune its traversal of the search space clearly helped it escape local optima. The multiple, iterative, adversarial cycles of counterresponses, anticipating criticisms, and testing different rebuttal strategies, strengthened the model's output by optimizing its searching abilities to move to a better location in the search space.

The value of revision and refinement is supported by the findings of (Madaan et al., 2023) and (Paul et al., 2024). Their work showed the efficacy of both iterative self-refinement and incorporating feedback from a critic model. Interestingly, (Madaan et al., 2023) explored the question of whether the value of refinement was due to the model's ability to generate multiple outputs – or stated in the language of search – whether it was benefiting from multiple search attempts. Controlling for this, their experimental results showed that model outperformance was driven specifically by the refinement process.

(Paul et al., 2024) elevate the granularity and sophistication of generating feedback – adopting the approach of training a critic model to provide feedback on erroneous intermediate steps. A direct comparison of performance on benchmarks between the critic model approach and the self-refinement framework of (Madaan et al., 2023) which relied on conventional prompting showed that the critic model delivered stronger results across the board. This highlights the value of a "specialized critic" that is more directly trained on the task.

## 5.4 Model Enhancement – Unleashing the Power of AI Engineering

In the absence of axiomatic truths and limited analytical rigor, experimental research data offers useful insights when trying to adjust and improve the architecture of models. The research literature is clear that structure, specialization, and sequential progression provide a blueprint for agent framework design. Refinement is a vivid illustration of this – it provides a final layer of enhancement to a structured process to remedy deficiencies and capture missed value. When refinement is expressed through more specialized architectural features (e.g., critic models), the effect is even more pronounced.

This sentiment has permeated machine learning discourse. Reasoning chains are increasingly being deconstructed into progressive steps like planning, execution, monitoring, and refinement. These steps constitute a strategic framework that can guide agents in open-world environments. This can be seen in the work of (Wang et al., 2023) who deploy an interactive planning approach rooted in goal decomposition and which integrates self-explanation of feedback.

This was borne out by the design and results of our experiment: creating two multi-agent pipelines with specialized agents performing discrete tasks and the progressive transformation of intermediate outputs along the chain of agents leading to a final robust output. Both models produced strong, coherent, organized, and compelling responses. The outperformance of the 8-agent recursive complex model over the simple 4-agent model is further evidence of the additive value of deeper specialization and structure as expressed by the recursive refinement architecture. Overall, our decomposition approach of modularizing each agent and linking them in a pipeline structure was a source of deep strength and stability for model performance.

More recent research decisively calls for treating AI development as an engineering discipline. Applying the rigor and discipline of software engineering principles to build robust, verifiable, and accurate models is seen as crucially important (Neary, 2024). This notion of "compositional" AI systems suggests shifting from a monolithic system to a collection of modules subject to independent and individual development and testing. It will be interesting to see how much value can be extracted from this enterprise before diminishing marginal returns or fundamental architectural limits on representational capacity start to materialize.

The continued and ongoing evolution of multi-agent frameworks with specialized agents performing specific functions, and the decomposition of reasoning into discrete steps (planning, execution, etc.), demonstrates the rising value of structure, specialization, and sequential progression in optimizing the performance of language models.

Decomposition raises important questions – is more decomposition always better? What is the right level of granularity for modularity? At what point do we encounter diminishing marginal returns? The work of (Khot et al., 2023) provides a fascinating starting point for addressing such questions. Their research design decomposes tasks into sub-tasks and allocates these to sub-task-specific LLMs each having their own few-shot prompts. They found that this led to superior performance compared to standard few-shot prompting.

Similar efforts to extend LLM capabilities have focused on providing the model with access to external tools such as calculators, search engines, programming interpreters, web browsers, and API calls. Interesting examples of this are found in the work of (Chen et al., 2023) and (Schick et al., 2023). Eschewing traditional prompting, (Chen et al., 2023) employ "program-of-thought" prompting in which computational steps are delegated to an external interpreter so that reasoning steps can be expressed as Python programs. (Schick et al., 2023) introduce a model with self-supervision that is trained to decide when and which APIs to call and how to synthesize this in its final generated output. Each approach outperformed baseline models.

## 6.0 Limitations and Future Work

### 6.1 Limitations

Several limitations must be acknowledged in assessing this paper. Our study used a small sample size of 9 data points. A next step to assess generalizability and statistical significance would be to use a larger pool of questions. Future research should focus on extending the GIS search architecture to other domains such as mathematics or programing to determine cross-domain applicability. While all results were generated using a single LLM (Claude 3 Sonnet), a worthwhile comparative analysis would be to re-run this experiment across different foundational models to identify model discrepancies in performance. Concerning the RAG database, further investigation is warranted into the impact of RAG corpus size and curation on the recursive refinement layer's effectiveness.

There is also an irreducible degree of subjectivity in using an LLM agent to evaluate and judge the quality of arguments given the inherent stochastic, probabilistic nature and inherent limitations of language models. A broader study would consider the use of an ensemble of arbiters to normalize for potential biases like verbosity or positional preference. To build on these findings, a future validation must employ a double-blind, multi-rater methodology with external domain experts to eliminate confirmation bias.

This study only compares two linear pipeline architectures. The outperformance of the complex pipeline validates our thesis, but a more comprehensive test would compare a structured linear pipeline against a decentralized, swarm-based architecture perhaps also incorporating RR.

More broadly, given that language model explainability and interpretability are in an evolving and speculative state, our pronouncements, explanations, and proposed search traversal framework (gradual, incremental, and sequential) cannot be rigorously established. Further lines of research however may provide the formal verification mechanisms that ascertain – to varying degrees – some of the concepts advanced.

### 6.2 Future Directions for Research

While the results surveyed in the broader research literature are impressive, some caveats are in order. The problems used for academic experiments are often narrow, well-defined, and contrived questions that have the artificial quality of being research constructs. It is hard to know whether these results translate well to open-ended, vague, and amorphous tasks that defy easy categorization or representation. Often just articulating and defining the problem is deeply challenging. Having noted that model generation is extremely sensitive to small variations in prompts, this is therefore far from a trivial concern.

For instance, it is not obvious how a complex and multifactorial objective like "extract the maximum amount of valuable minerals from the ocean floor while minimizing disruption to marine life" can be decomposed and modularized in a formulaic way. Outsourcing this to an LLM by asking it to "generate a plan" and then to "prioritize sub-goals" leads us to a world of conjecture and uncertainty. For instance, is it sufficient to instruct the model to "develop a plan" given that not all plans are created equal? How do we influence the planning process? What is the best way for adjusting and managing strategic divergences from an initial plan?

Questions around decomposition and hierarchy soon extend to discussions around the expansion of capabilities. What sort of external databases, libraries, and tools should the LLM have? Should natural language be substituted for programmatic prompting? How much additional supervision should we apply in the form of training models on erroneous reasoning, domain primitives, or other useful constructs?

Consider a mathematical problem solving model that is given access to a library of problem-solving heuristics (e.g., invariance, extremal principle, graphs, etc.), proof styles (proof by contradiction, induction, constructive, etc.), theoretical constructs like leveraging techniques from one domain to solve problems in another (e.g., the use of real analysis in number theory), or even a natural language catalog of the motivations and intuitions of major mathematical discoveries. There are an arbitrary number of augmentations, compositions, and abstraction which can be applied in service to enhancement. The dust has not yet settled in finding canonical design patterns.

Perhaps the ultimate pursuit (excepting artificial general intelligence) in this space is the development of AI systems that can optimize the design of such architectural frameworks. Stated otherwise, building AI that builds AI. Our analysis has relied heavily on LLMs, the dominant architecture, but the next advance forward might require a new paradigm shift or transition to a new family of models. Although we have now firmly entered the realm of conjecture, the questions generated by AI reasoning and the supporting architectural and design considerations will remain enduring and relevant.

## 7.0 Conclusion

We set out to determine the efficacy of a multi-agent pipeline integrating recursive refinement in improving language model performance leveraging a GIS-search framework. Implementing a comparison between two multi-agent pipelines – one without and one with a recursive refinement layer – we ran a simulation of founding father agent personas generating responses to a range of questions. The inclusion of recursive refinement led to improvement in model performance – the overall arguments produced were more compelling and persuasive.

This result supports our central thesis: that optimizing LLM computation and reasoning calls for structuring a multi-agent pipeline to ensure a traversal of the search space that is gradual, incremental, and sequential. This principle is best expressed by architectures that seek to modularize, decompose, and arrange critical steps in a structured and sequential progression.

Ultimately, the design of AI architectures will increasingly leverage principles of specialization and decomposition, and recursive refinement will play a central role in enhancing and extending the reasoning and learning capabilities of models. If software engineering is the art of managing complexity, then AI engineering may be construed as the art of managing the complexity of representational learning.

## 8.0 Code and Data Availability

The complete code base and implementation of the multi-agent frameworks including all agent classes and architectures, experimental scripts, and evaluation code and criteria, and RAG system is publicly available at:

https://github.com/alvco/Founding_Fathers_AI

# 10.0 Appendix

The table below lays out the Arbiter Agent scores for all models, agents, and questions:

| Q. ID | Topic | Agent | Model Type | Final Score | Structure Score | Depth Score | Support Score | Rhetoric Score |
|---|---|---|---|---|---|---|---|---|
| 1 | Arms Sales | Hamilton | Simple | 75 | 8 | 7 | 7 | 8 |
| 1 | Arms Sales | Hamilton | Complex | 82.5 | 8 | 8 | 8 | 9 |
| 1 | Arms Sales | Jefferson | Simple | 65 | 7 | 6 | 6 | 7 |
| 1 | Arms Sales | Jefferson | Complex | 87.5 | 9 | 9 | 8 | 9 |
| 1 | Arms Sales | Madison | Simple | 82.5 | 8 | 8 | 8 | 9 |
| 1 | Arms Sales | Madison | Complex | 90 | 9 | 9 | 9 | 9 |
| 2 | Immigration | Hamilton | Simple | 72.5 | 7 | 7 | 7 | 8 |
| 2 | Immigration | Hamilton | Complex | 90 | 9 | 9 | 9 | 9 |
| 2 | Immigration | Jefferson | Simple | 67.5 | 7 | 6 | 6 | 8 |
| 2 | Immigration | Jefferson | Complex | 85 | 9 | 9 | 8 | 8 |
| 2 | Immigration | Madison | Simple | 72.5 | 7 | 7 | 7 | 8 |
| 2 | Immigration | Madison | Complex | 90 | 9 | 9 | 9 | 9 |
| 3 | Annexation | Hamilton | Simple | 67.5 | 7 | 7 | 6 | 7 |
| 3 | Annexation | Hamilton | Complex | 90 | 9 | 9 | 9 | 9 |
| 3 | Annexation | Jefferson | Simple | 67.5 | 7 | 6 | 7 | 7 |
| 3 | Annexation | Jefferson | Complex | 90 | 9 | 9 | 9 | 9 |
| 3 | Annexation | Madison | Simple | 75 | 8 | 7 | 8 | 7 |
| 3 | Annexation | Madison | Complex | 90 | 9 | 9 | 9 | 9 |

The responses below were provided by the Arbiter Agent in its evaluation of all the outputted responses across all models, agents, and questions. In each case "Argument B" refers to the output of the complex model and "Argument A" refers to the output of the simple model:

Question 1 – Hamilton:

While both arguments are well-structured and rhetorically compelling, Argument B demonstrates superior depth by adding an economic/industrial dimension to the strategic analysis that Argument A lacks. Argument A presents a straightforward realpolitik case grounded in the Jay Treaty precedent, effectively arguing that national interest must trump moral sentiment. However, Argument B enriches this framework by connecting arms sales not merely to immediate strategic advantage but to long-term industrial capacity and economic sovereignty, referencing the Report on Manufactures as theoretical foundation. Argument B also more directly engages with the primary counterargument (risk of blowback), acknowledging it before dismissing it, whereas Argument A more summarily rejects moral concerns. Rhetorically, both are strong, but Argument B's concluding triadic structure and memorable phrases like 'Let other nations debate...while their forges grow cold' give it a slight edge. The combination of broader analytical scope, more nuanced engagement with objections, and marginally superior rhetoric results in Argument B's victory with a final score of 82.5 versus 75.

Question 1 – Jefferson:

Argument B significantly outperforms Argument A across all four criteria. Structurally, Argument B is more sophisticated, explicitly acknowledging and addressing counterarguments rather than simply asserting principles. In terms of depth, Argument B explores multiple dimensions of the issue—moral precedent, institutional corruption, international perception, and long-term consequences—while Argument

A remains more surface-level in its treatment. For support and justification, Argument B provides more complete reasoning chains and addresses the opposition's strongest points, whereas Argument A relies more heavily on assertion. Rhetorically, Argument B employs varied strategies including concession, rhetorical questions, and powerful parallel constructions, making it more persuasive than Argument A's more straightforward elevated rhetoric. The 22.5-point difference in final scores (87.5 vs 65) reflects Argument B's consistent superiority across all evaluated dimensions.

Question 1 – Madison:

While both arguments are excellent and make fundamentally similar principled cases, Argument B edges ahead across all four criteria. Structurally, B more effectively acknowledges the costs and complexity of its position before making its case, which strengthens rather than weakens its logical force. In terms of depth, B makes the crucial insight more explicit that foreign and domestic policy are unified by a single principle, not separate questions requiring different standards. For support and justification, B more directly engages with counterarguments ('I do not dispute...nor do I dismiss') and offers specific alternatives, demonstrating intellectual honesty that enhances credibility. Rhetorically, both are powerful, but B's acknowledgment of real trade-offs before asserting principle makes it more persuasive to skeptical audiences. The key differentiator is that Argument B demonstrates slightly more sophistication in acknowledging the genuine costs of its principled position while still maintaining that stance, making it both more intellectually rigorous and more practically persuasive. The final scores of 82.5 versus 90 reflect consistent superiority across all weighted criteria.

Question 2 – Hamilton:

Argument B surpasses Argument A across all four criteria. Structurally, B employs a more sophisticated dialectical approach that anticipates and addresses counterarguments, particularly regarding executive discretion, while A follows a more linear advocacy structure. In depth, B explores tensions and trade-offs explicitly (flexibility vs. rigidity, discretion vs. abuse), acknowledges the impossibility of perfect systems, and provides multi-layered analysis of institutional frameworks, whereas A presents a more straightforward case without engaging complexities. For support and justification, B provides concrete scenarios (European manufacturers, economic depression), detailed institutional mechanisms (legislative supremacy, judicial review, electoral accountability), and specific examples of when flexibility is needed, while A relies more on abstract assertions and a single historical analogy. Rhetorically, B demonstrates greater variety through effective use of rhetorical questions, builds momentum through extended argumentation, and employs more dynamic pacing, while A, though maintaining strong Hamiltonian voice, is somewhat repetitive in its core message. The decisive advantage lies in B's willingness to engage with the complexity of implementation and potential objections, making it a more complete and persuasive argument.

Question 2 – Jefferson:

While both arguments eloquently invoke Revolutionary principles and draw parallels between selective immigration and aristocratic privilege, Argument B demonstrates significantly superior argumentation. The decisive advantage lies in Argument B's explicit acknowledgment and engagement with the most obvious counterargument: that political communities possess sovereign authority over territorial boundaries and that there exists a theoretical distinction between natural rights and membership criteria. Rather than ignoring this objection, Argument B confronts it directly and explains why the dangers of selective policy outweigh any benefits from exercising sovereign discretion. This intellectual honesty strengthens rather than weakens the argument. Argument B also provides more sophisticated analysis through its discussion of how artificial distinctions evolve, its risk-benefit framework, and its invocation of Paine as additional authority. Structurally, Argument B is more complete, moving from thesis to counterargument to refutation to conclusion in a tighter logical progression. While Argument A scores well on rhetoric (8/10), it falls short on depth (6/10) and support (6/10) because it presents a one-dimensional analysis that fails to grapple with the complexity of distinguishing between citizens and non-citizens. Argument B achieves higher scores across all categories (9, 9, 8, 8) by demonstrating that one can acknowledge nuance while maintaining a principled position, resulting in a final score of 85 versus 67.5.

Question 2 – Madsion:

Argument B surpasses Argument A across all four criteria. Structurally, it employs a more sophisticated dialectical approach that acknowledges opposing concerns before reframing the debate, whereas Argument A follows a more linear progression. In terms of depth, Argument B demonstrates greater nuance by engaging with the complexity of balancing competing interests (growth vs. cohesion, diversity vs. unity) and proposing a middle-ground solution, while Argument A takes a more one-dimensional stance prioritizing existing citizens. For support and justification, Argument B provides more comprehensive reasoning directly grounded in Federalist principles (extended republic theory, faction multiplication) that are more relevant to the immigration question than Argument A's debt assumption analogy. Rhetorically, Argument B is more sophisticated in its use of strategic concession and reframing, building credibility before presenting its case, while Argument A, though powerful, is more straightforwardly assertive. Most significantly, Argument B offers specific, actionable policy recommendations with clear rationale, demonstrating practical wisdom alongside theoretical sophistication. The 17.5-point difference in final scores (90 vs. 72.5) reflects Argument B's consistent superiority across all evaluated dimensions.

Question 3 – Hamilton:

Argument B significantly outperforms Argument A across all four criteria. Structurally, Argument B presents a clear framework distinguishing ultimate objectives from immediate methods and systematically addresses counterarguments, while Argument A tends toward repetition without progressive development. In depth, Argument B engages with multiple levels of consent, distinguishes between types of influence, and grapples with moral complexities more thoroughly than Argument A's somewhat abstract principles. For support and justification, Argument B provides a detailed four-point strategy, systematically compares three alternatives, and offers extensive reasoning for each claim, whereas Argument A's criteria for justified expansion remain somewhat vague. Rhetorically, Argument B's strategic concessions ('Let me speak with the candor this question demands') and systematic refutation of objections prove more persuasive than Argument A's elevated but repetitive language. The 22.5-point margin (90 vs 67.5) reflects Argument B's comprehensive superiority in presenting a complete strategic framework with specific mechanisms while engaging more deeply with the practical and moral dimensions of the question.

Question 3 – Jefferson:

Argument B significantly outperforms Argument A across all four criteria. In structure, B anticipates and addresses the most powerful counterargument (Madison's extended republic theory), while A follows a more linear progression without engaging potential objections. In depth, B makes sophisticated theoretical distinctions between governing consenting citizens across distance versus governing unwilling foreign populations, while A treats the issue more superficially. For support and justification, B provides rigorous reasoning by explicitly addressing why Madison's theory doesn't warrant the proposed expansion, noting it 'presumes a foundation of shared principles, common language, and voluntary union—the very conditions that would be violated by forcible annexation.' A offers sound but straightforward reasoning without this level of analytical rigor. Rhetorically, both arguments are strong, but B combines powerful language with intellectual sophistication, respectfully engaging with 'my esteemed friend Mr. Madison' while firmly rejecting the application of his theory. The critical distinction B draws—'it is the difference between self-government and subjugation'—encapsulates a more nuanced understanding of republican principles than A's more conventional appeals to Revolutionary precedent.

Question 3 – Madison

Argument B excels across all four criteria. In structure, it employs a more sophisticated organization by immediately acknowledging potential counterarguments and turning them into support for its thesis. In depth, it demonstrates exceptional nuance by grappling with the tension between principle and practice regarding Indian removal, showing intellectual honesty that strengthens rather than weakens its position. For support and justification, it uses an effective a fortiori argument (if we failed with smaller populations, we cannot succeed with larger ones) combined with appeals to consent, constitutional theory, and practical experience. In rhetoric and style, it achieves greater emotional resonance and authenticity through personal vulnerability ('I have myself harbored private reservations') and powerful rhetorical questions. While Argument A is well-constructed and makes sound points about republican government, Montesquieu, and Federalist No. 10, it lacks the moral complexity, self-awareness, and rhetorical power that distinguish Argument B. The willingness to acknowledge past failures and use them productively represents sophisticated political philosophy that elevates Argument B significantly above its competitor.